# D-GAN: Deep Generative Adversarial Nets for Spatio-Temporal Prediction

Divya Saxena
University Research Facility in Big Data Analytics,
The Hong Kong Polytechnic University,
Kowloon, Hong Kong
divya.saxena@polyu.edu.hk

Jiannong Cao
Department of Computing and
University Research Facility in Big Data Analytics,
The Hong Kong Polytechnic University,
Kowloon, Hong Kong
csjcao@comp.polyu.edu.hk

*Abstract*— Spatio-temporal (ST) data for urban applications, such as taxi demand, traffic flow, regional rainfall is inherently stochastic and unpredictable. Recently, deep learning based ST prediction models are proposed to learn the ST characteristics of data. However, it is still very challenging (1) to adequately learn the complex and non-linear ST relationships; (2) to model the high variations in the ST data volumes as it is inherently dynamic, changing over time (i.e., irregular) and highly influenced by many external factors, such as adverse weather, accidents, traffic control, PoI, etc.; and (3) as there can be many complicated external factors that can affect the accuracy and it is impossible to list them explicitly. To handle the aforementioned issues, in this paper, we propose a novel deep generative adversarial network based model (named, D-GAN) for more accurate ST prediction by implicitly learning ST feature representations in an unsupervised manner. D-GAN adopts a GAN-based structure and jointly learns generation and variational inference of data. More specifically, D-GAN consists of two major parts: (1) a deep ST feature learning network to model the ST correlations and semantic variations, and underlying factors of variations and irregularity in the data through the implicit distribution modelling; (2) a fusion module to incorporate external factors for reaching a better inference. To the best our knowledge, no prior work studies ST prediction problem via deep implicit generative model and in an unsupervised manner. Extensive experiments performed on two real-world datasets show that D-GAN achieves more accurate results than traditional as well as deep learning based ST prediction methods.

## I. Introduction

In recent times, developing efficient urban applications using spatio-temporal (ST) data, such as air and water quality forecasting [1], crowd flows prediction [2], cellular traffic prediction [3], demand prediction [4] have drawn increasing attention. These applications require real-time analysis and accurate prediction capabilities. For example, an accurate taxi demand prediction model can help drivers to increase their revenue by finding passengers easily and reduce the overall on-road traffic and carbon footprints in a city.

Recently, deep learning based models have been proposed to learn the complex and non-linear prediction functions and performed well w.r.to model accuracy in various research areas, such as computer vision and natural language processing, gaming, etc. Recent studies in ST prediction have explored these deep learning models and have modeled a city/locations, its areas and data value into matrices analogous to the pixel matrices of an image, i.e., ST map [5][6]. In this case, for the given sequences of ST maps/images/matrices, model tries to predict the next ST map accurately. Researchers have proposed Convolutional neural network (CNN), Recurrent neural network (RNN) and RNN-CNN based framework [4][7][8] to model the complex spatial correlation and temporal dependencies. These models achieved high accuracy in comparison to traditional time-series prediction approaches.

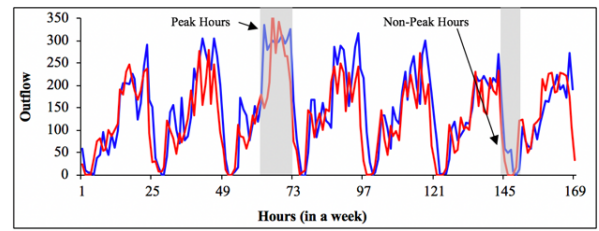

Fig. 1. High variation in the ST data volume, e.g., taxi demand variation pattern for two consecutive weeks in an area of New York City

Existing deep learning based ST prediction models have proposed to learn the ST characteristics of data. However, it is still very challenging (1) to adequately learn the complex and non-linear ST relationships; and (2) to model the high variations in the ST data volumes as it is inherently dynamic, changing over time (i.e., irregular/stochastic) (see Fig. 1) and highly influenced by many external factors, such as adverse weather, accidents, traffic control, PoI, etc.

On the other hand, instead of spatial dependency and temporal dynamics challenges to model ST correlations more accurately, the existing models also have three major limitations: (1) when a model trained using the mean squared error (MSE) *by construction* between the ground truth and the predicted data, model chooses an average over many slightly different data. This causes the prediction of low accuracy [9]. Existing deep learning based ST prediction approaches use deterministic models, such as feed-forward and RNN which assume that outcome is deterministic and cannot capture the stochastic behavior in the data [9][27], which is utmost important factor for an accurate ST data prediction model. Therefore, a highly accurate ST prediction approach requires a model which can work with multi-modal outputs; (2) existing ST prediction models use multiple events data during the training to improve the model's accuracy. There can be many complicated factors that can affect the accuracy and it is impossible to list them explicitly, therefore, ST prediction approach requires an efficient and fine-grained deep ST feature learning model which can learn the data distribution implicitly; (3) existing approaches lack an efficient inference mechanism to support the reasoning about data at an abstract level. This is important for policy makers to identify what contributes to the model's improvement.

To address the above challenges and limitations, we attempt to handle ST data prediction in a more efficient way. Our idea is conceptually inspired by a deep implicit generative model, Generative Adversarial Networks (GAN) [9] which generates the raw pixels of future images from a sequence of context images. GAN works with multi-modal outputs and learns rich distributions implicitly over images, audio, and data which are hard to model with an explicit likelihood.

In recent times, GAN has shown a great potential and has been applied to various research areas successfully, such as image generation [10][11], video prediction [12], etc. However, applying image generation and video prediction approaches directly are not applicable for predicting ST data of urban applications, such as taxi demand, crowd flow, etc. An image generation process can consider the large changes in appearance between the input and output image, but it cannot adequately handle spatial variations. On the other hand, video prediction process can take spatial changes into account, but appearance remains largely the same from image-to-image. In addition, the prediction of next image in image sequences is highly dependent on its previous image, i.e., temporal dynamics are not considered.

Moreover, we find that directly applying basic GAN to ST data prediction problem is less effective, because: (1) GAN model with input noise can indeed generate better results, but fails to adequately cover the space of possible futures; (2) how to leverage additional information, such as weather, PoI, with the model is still untouched in existing models; (3) basic GAN does not support an inference network which is required to better understand what contributes to the model's improvement.

In this paper, we propose a novel deep generative adversarial network based model (named, D-GAN) for more accurate ST data prediction. D-GAN learns to predict future ST map from a sequence of ST maps through learning construction of non-trivial internal representation accurately. The main aim to use feature learning is that features from different spaces are complementary for more accurate prediction. D-GAN combines the variational inference with GAN into an unsupervised generative model to model stochastic behavior in the ST data. GAN does not use element-wise similarity measure by construction which gives GANs an ability to predict more accurately. In training time, an inference network estimates the distribution of latent variables, while a generative network learns to reconstruct input space from the latent variables.

In addition, accurate modeling of sequence of ST maps involves high-level understanding of highly structured data which requires the learning of an internal representation to capture complex ST correlations and underlying factors of variations. We also exploit the ability of ConvLSTM [13] and 3D-ConvNet structures to model long-term trends and local spatial dependencies in the ST data, respectively. We use ConvLSTM with GAN structure as ConvLSTM alone cannot handle variations in the ST data explicitly.

In this paper, we take a case study of taxi demand prediction in a city and evaluate D-GAN on two real-world NYC Taxi and Bike datasets. Experiment results show that D-GAN outperforms 14 mainstream baselines w.r.to root mean squared error and mean absolute error performance metric. D-GAN achieves a more realistic prediction map due to learning ST correlations and high data variations implicitly. To the best of our knowledge, no prior work studies ST prediction via deep implicit generative model and in an unsupervised manner.

In summary, our contributions are summarized as follows:
- We propose a novel deep generative adversarial network based model (named, D-GAN) to deeply capture the underlying ST data distribution implicitly for more accurate ST prediction. Different from the existing deep models for ST prediction, D-GAN combines GAN and VAE and jointly learns generation and variational inference of ST data in an unsupervised manner. This, to the best of our knowledge, has not been done in prior ST prediction research.
- We design a general fusion module to fuse heterogeneous multiple data sources, such as weather data, time information, PoIs data, etc., for reaching a better inference. The learned latent representations are merged with the ST data representation as external factors have significant impact on the variation of ST data volume. D-GAN is highly flexible and extendable as it can be easily extended for a new ST dataset with multiple data sources.
- Results of a case study on two real-world benchmark datasets show that D-GAN achieves more accurate results than traditional as well as deep learning based ST prediction methods.

II. BACKGROUND

Our D-GAN model uses the GAN framework to generate the next ST map from a sequence of given ST maps. GANs use the concept of a non-cooperative game in which two networks, a generator (G) and a discriminator (D), are trained to play against each other.

The goal of $G$ is to generate samples resembling to samples generated from true data distribution, while $D$'s purpose is to distinguish between samples drawn from $G$ and samples drawn from the true data distribution. $D$ assigns higher probabilities to real data samples and lower probabilities to samples generated by $G$, where GAN training simultaneously keeps trying to move the generated samples towards the real data manifolds using the gradient information provided by the $D$. The data $x$ drawn from the true data distribution, $p_{data}$, noise vector $z$ comes from a prior distribution $p_z$ and $p_g$ is the $G$'s distribution over data $x$. GAN considers a Gaussion prior distribution with zero-mean and unit variance. $G$ takes latent vector $z$ as input and outputs an sample G($z$) with the goal of bringing G($z$) as close as possible to D($x$), i.e., $G$ generates samples that are hard for the $D$ to distinguish from real data. At the same time, the $D$ tries to avoid getting fooled by the generative model $G$. On the other hand, $D$ is simply a classifier in which D($x$) = 1 if $x \sim p_{data}$ and D($x$) = 0 if $x \sim p_g$, i.e., $x$ is from $p_{data}$ or from $p_g$. A minimax objective is used to train both $G$ and $D$ models jointly via solving:

$$\min_{\theta_G} \max_{\theta_D} V(G,D) = \min_G \max_D \mathbb{E}_{x \sim p_{data}}[\log D(x)] + \mathbb{E}_{z \sim p_z}[\log(1 - D(G(z)))] \quad (1)$$

$V(G, D)$ is a binary cross entropy function that is commonly used in binary classification problems [9]. Both $G$ and $D$ are trained by backpropagating the loss in Eq. 1 through their respective models to update the parameters.

Due to the two different objectives in Eq. 1, the update rule is defined as follows:

$$\{\theta_D^{t+1}, \theta_G^{t+1}\} \leftarrow \begin{cases} \text{Update} & \text{if } D(x) \text{ predicts wrong} \\ \text{Update} & \text{if } D(G(z)) \text{ predicts wrong} \\ \text{Update} & \text{if } D(G(z)) \text{ predicts correct} \end{cases}$$

where $\theta_D^{t+1}$ and $\theta_G^{t+1}$ are the parameters of $D$ and $G$, respectively, and $t$ is the iteration number. With the enough capability at $D$ and $G$, and sufficient training iterations, $G$ will be able to transform a simple prior distribution $p_g$ to more complex distributions, i.e., $p_g$ converges to $p_{data}$, such as $p_g = p_{data}$. A GAN model is well trained when equilibrium is achieved between $D$ and $G$, and $D$ cannot distinguish whether a sample is generated by the $G$ or generated from the real data distribution. In practice, the players are represented with deep neural nets and updates are made in parameter space.

In recent times, GAN has been introduced as an innovative technique for learning generative models of complex data distributions from samples. Instead of its remarkable success in generating realistic images, training a GAN is really challenging due to the various problems, such as *difficulty converging*, *mode collapse*, and *vanishing gradients* [9].

In addition, we use the Variational Autoencoders (VAEs) [14] with the GAN to learn the complex data distribution in an unsupervised manner. VAEs support both generation and inference by learning a bidirectional mapping between a complex data distribution and simple prior distribution. An autoencoder is a member of neural network models which learns compressed latent variables of a high-dimensional data. VAE is one of the autoencoders based on the Variational Bayesian and graphical model concept. VAE maps the input into a distribution rather than into a fixed vector.

Autoencoders learn the relationship between data and its latent code directly (i.e., explicit), while GAN learns to generate samples indirectly (i.e., implicit).

III. PROBLEM STATEMENT

**Definition 1 (Region).** *There are many definitions to define a location w.r.to different granularities and semantic meanings. In this study, we divide a city based on longitude and latitude into two-dimensional grid map of $m \times n$ size where a grid represents a region and a grid map named as ST map.*

**Definition 2 (Measurements).** *There are different types of measurements for a region that can be used for various ST applications, such as taxi demand and crowd flows prediction, and air quality prediction. In this study, we take a case study of on-demand service prediction in a city. Therefore, we use the demand values as measurements. For a region $R(m, n)$ and at a timestamp $t^i$ where $i > 0$, demand value is represented as $d_{t^i}^{R(m,n)}$.*

**Problem 1.** *Given the historical data $d_t$ for time interval $t = 0, ...., t-1$, our aim is to predict $d_{t+1}$ with high prediction accuracy.*

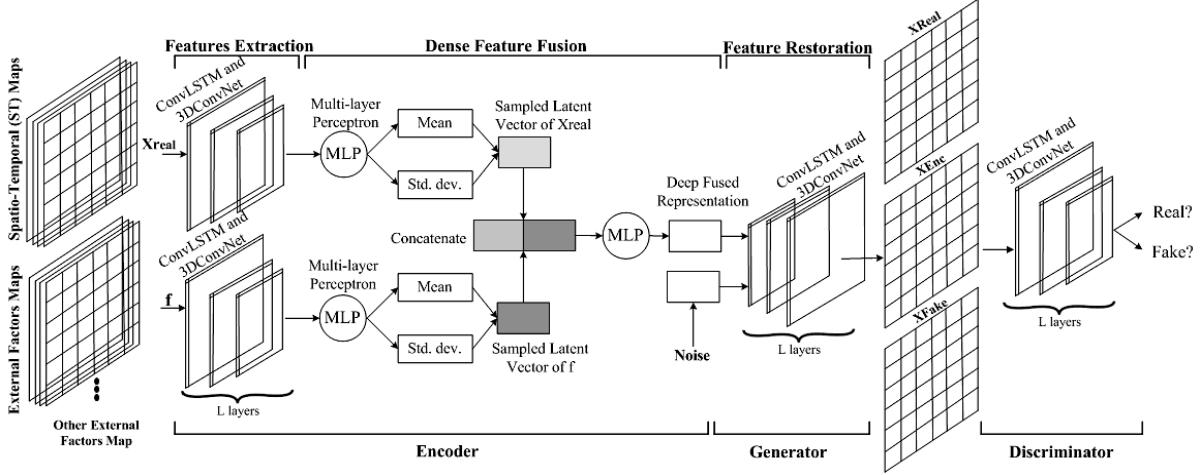

Fig. 2. D-GAN architecture

IV. D-GAN FRAMEWORK

In this section, we introduce the framework of D-GAN in detail whose overall network architecture is shown in Fig. 2.

*A. Network Architecture*

D-GAN comprises of four main components: *Encoder network* (E), *Generator/Decoder network* (G), *Discriminator network* (D) and *external factors fusion* (F). Encoder network $E$ is comprising of a probabilistic encoder which encodes the data space $x$ into latent code $x_c$. The inference network outputs parameters to the distribution $q(x_c|x)$. Generator network $G$ comprises of a probabilistic decoder that learns to reconstruct input space $x$ from representation $x_c$. The generative network outputs parameters to the likelihood distribution $p(x|x_c)$. $G$ network is trained using the adversarial process in which $G$ (i.e., decoder) learns to approximate the distribution of real data, while the discriminative network $D$ discriminate between samples from the real distribution and samples generated by $G$. In this study, $D$ learns to discriminate jointly in data and latent space for stabilized training and better learning. During the training, D-GAN uses a reconstruction loss between the $q(x_c|x)$ and $p(x|x_c)$ in addition to calculating the adversarial loss of $G$ and $D$ in backpropagation. Then, we design a general fusion network to integrate external factors from different domains with the data. The data of fusion module fed into $E$ and the generated latent code $f_c$ of external factors $f$ is merged with the extracted features $x_c$.

In D-GAN, $E$, $G$ and $D$ use a stack of ConvLSTM and 3D-ConvNet elements where ConvLSTM neural network is able to capture long-term trends in the ST map sequences, i.e., ST correlations while 3D-ConvNet captures local spatial dependencies. ConvLSTM has convolutional structures in both the input-to-state and state-to-state transitions. The 3D convolution has ability to model the cross-temporal traffic correlations. 3D-ConvNets can capture better short-term data volume fluctuations and improves overall model's generalization abilities as it maintains the relationship between neighboring input points by sharing weights across different locations in the input and ST locality in feature representations.

***Encoder and Generator***: In D-GAN, encoder first processes the real data by multiple stack of ConvLSTM and 3D-ConvNet, and a multi-layer perceptron (MLP) to produce a condensed feature vector $FV_x$, i.e.,

$$FV_x = \text{MLP}\left(\text{ConvNet3D}(\ldots \text{ConvLSTM}(x_{real}))\right),$$
$$= \text{MLP}\left(\text{ConvNet3D}\left(\text{ConvLSTM}^L(x_{real})\right)\right), \quad (2)$$

where $x_{Real}$ is a real data space, $FV_x$ is the extracted feature vector of $x_{Real}$, and $L$ is the number of ConvLSTM layers. Then, we use a Variational Bayesian method with the multivariate Gaussian assumption and variational lower bound loss function where model calculates the mean ($\mu$) and variance ($\sigma$) of the distribution explicitly using the *reparameterization trick* as follows:

$$FV^r_{x_{real}} = \mu + \sigma \odot \epsilon, \text{where } \epsilon \sim \mathcal{N}(0, I), \quad (3)$$

where $\epsilon$ is an auxiliary independent random variable and $\odot$ is the element-wise product. We use the Kullback-Leibler (KL) divergence as regularization term to ensure that $q(FV^r_{x_{real}} | x_{real})$ does not diverge too much from prior $p(FV^r_{x_{real}})$. The same procedure is performed for the external factors $f$ to incorporate them into the prediction model. D-GAN supports a general fusion network to integrate external factors from different domains with the model. We design a feature extraction module with similar stack of ConvLSTM layers, a ConvNet3D and an MLP. The learned auxiliary feature vector with Gaussian assumption is represented as $FV^r_f$. The extracted two condensed feature vectors are concatenated together, i.e., $FV_{cat} = [FV^r_{x_{real}}, FV^r_f]$, and fed into the *Decoder* (i.e., G), i.e., an MLP and stacked ConvLSTM and 3D-ConvNet layers to reconstruct the input ST map in original size:

$$x_{Enc} = \text{ConvNet3D}\left(\text{ConvLSTM}^L(\text{MLP}(FV_{cat}))\right), \quad (4)$$

where $x_{Enc}$ is the reconstructed ST map of $x_{real}$. Furthermore, a noise vector $z$ is passed to the $G$ as input for generating the reconstructed ST map of noise vector $x_{Fake}$.

***Discriminator:*** $D$ learns to determine whether a generated ST map is from the ground truth or produced by $G$. In D-GAN, we concatenate the latent code $FV$ with its generated ST map $x_{Enc}$ to jointly learn the latent code and data space as we notice it supports fast convergence, better learning and high training stability, i.e., $y_{Enc} = [x_{Enc}, FV_{cat}]$. We also generate $y_{Fake} = [x_{Fake}, FV^*_{cat}]$ and $y_{real} = [x_{real}, FV_{cat}]$ where $FV^*_{cat}$ is the noise feature vectors. Then, we use the similar stacked ConvLSTM layers and a ConvNet3D layer for the implementation of $D$:

$$D_{out} = \sigma\left(\text{ConvNet3D}\left(\text{ConvLSTM}^L(y)\right)\right), \quad (5)$$

where $y$ is either the generated encoded ST map $y_{Enc}$ with latent code or the ground truth $y_{real}$ with latent code or the generated fake ST map $y_{Fake}$. $D_{out}$ is the predicted probability of input $y$ being real or fake. $G$ and $D$ network are trained simultaneously till $D$ cannot discriminate the ST maps generated by $G$ with ST maps generated from real data. After training, $G$ is used to generate samples similar to real data.

*B. Loss Functions*

In this section, we discuss loss functions used for the D-GAN. We categorize them into two parts: the adversarial losses and a reconstruction loss.

*1) Adversarial Losses* ($L_{adv}$): The adversarial losses are used to find the equilibrium between $G$ and $D$ during the adversarial training process. In D-GAN, the adversarial loss of $D$ ($L^D_{GAN}$) is as follows:

$$L^D_{GAN} = \|D(y_{real}) - 1\|^2_2 + \|D(y_{Fake}) - 0\|^2_2 + \|D(y_{Enc}) - 1\|^2_2, \quad (6)$$

We use least squares loss function instead of binary cross entropy used in basic GAN to evaluate the difference. On the other hand, $G$'s aim is to generate real-looking samples w.r.to $D$, so to minimize the $G$ loss ($L^G_{GAN}$), $G$ tries to reduce the difference between $D(y_{Enc})$ and the true label, and $D(y_{Fake})$ and the true label as shown in Eq 7.

$$L^G_{GAN} = \|D(y_{Fake}) - 1\|^2_2 + \|D(y_{Enc}) - 1\|^2_2, \quad (7)$$

*2) VAE Loss* ($L_{VAE}$): In our case, VAE loss comprises of two losses as follows:
- KL divergence ($D_{KL}$): It measures the divergence between two probability distributions.
- Reconstruction loss ($L_R$): It calculates element-wise deviations between the ground truth and the reconstructed ST map to find the local differences between grids. $L_R$ is defined as the element-wise L2-norm:

$$L_R = \frac{1}{mn}\|x_{real} - x_{Enc}\|_2,$$

$mn$ is the total number of regions in a ST map. D-GAN uses element-wise L2-norm as reconstruction loss with the adversarial loss due to nature of the ST data

Thus, the overall objective of D-GAN is as follows:

$$L_{D-GAN} = L_{VAE} + L^G_{GAN} + L^D_{GAN}$$

V. EXPERIMENTS

*A. Case Study*

We formulate the demand prediction as a ST prediction problem in which both the input and the prediction target are ST sequences. The main aim of this task is to learn an accurate model to predict the total number of requests for a particular service in each grid of ST map during each time slot where a time slot can be an hour, or a day, or a week. We use two large-scale datasets collected at NYC city: Yellow Taxi dataset (Taxi) [15] which contains requests from 01/01/2016 to 30/06/2016 and CitiBike trip dataset (Bike) [16] which contains requests from 01/01/2016 to 31/01/2016 for the demand prediction. We represent a city as ST map where $l_s$ = <$lat_s$, $long_s$> and $l_e$ = <$lat_e$, $long_e$> denote the start $s$ and end $e$ location coordinates of a city, respectively. The square area clustered into 9×9 non-overlapping regions which represents as a ST map. The demand is the number of Taxis/Bikes requests at a region in a time interval $t$. A day divided into 24 hours timeslots, i.e., a ST map represents demand in an hour. We use the external factors, such as Point of Interest (PoI), weather data and weekend/weekday with the historical data for demand prediction.

*B. Evaluation Metrics*

We use both Rooted Mean Square Error (RMSE) and Mean Absolute Error (MAE) as the evaluation metrics to evaluate the performance of D-GAN:

$$\text{RMSE} = \sqrt{\frac{1}{mn}\Sigma_i(x_i - \hat{x}_i)^2}, \qquad (8)$$

$$\text{MAE} = \frac{1}{mn}\Sigma_i(x_i - \hat{x}_i), \qquad (9)$$

where $\hat{x}_i$ and $x_i$ are the prediction value and real value of $i^{th}$ ST map, and $mn$ is total number of regions in a ST map.

*C. Baselines*

To illustrate the effectiveness of our model, we compare it with 14 mainstream baseline methods and tuned the parameters for all methods. The models used for the comparison are as follows:

- **Simple Moving Average:** It predicts the data value using average values of previous data values at the location given in the same relative time interval.
- **Weighted Moving Average:** It is the variant of the historical average model with a weight associated with observations.
- **Autoregressive integrated Moving Average (ARIMA):** It is a well-known model which combines moving average and autoregressive components for time-series data modeling.
- **Linear Regression (LR):** We use ordinary least squares regression (OLSR) model to estimate the relationship between multiple variables.
- **Gradient Boosting Decision Tree (XGBOOST):** It is an ensemble learning in which many ML models are trained at once for better performance.
- **Random Forest (RF):** It is a widely used ensemble method which achieves comparable performance with XGBOOST.
- **Multiple Layer Perceptron (MLP):** It is a neural network of four fully connected layers with hidden units 128, 128, 64, and 64.
- **Long Short Term Memory Neural Network (LSTM):** It is a neural network of a LSTM layer with a fully connected layer.
- **Convolutional Neural Network (CNN):** We use the two convolutional layers followed by MaxPooling layer and a fully connected layer.
- **ST-ResNet [2]:** It is the deep convolutional based residual networks which is used for the grid-based traffic flow prediction.
- **DMVST-Net [8]:** It is a deep multi-view ST neural network, i.e., temporal, spatial and semantic view, also used for grid-based prediction.
- **GAN-based LSTM (G-LSTM):** We use the basic GAN with LSTM for the grid based prediction. The numbers of hidden units are 256, 512, and 1024.
- **GAN-based CNN (G-CNN):** We use the basic GAN with two convolutional and one fully connected layer.
- **GAN-based ConvLSTM (G-ConvLSTM):** We build an end-to-end trainable model using the basic structure of GAN with four ConvLSTM and one 3DConvNet with the same parameters as D-GAN.

The number of regions used for LSTM, CNN and MLP are 30 for Taxi dataset while 117 for Bike dataset for best performance.

*D. Preprocessing and Parameters*

We use the Min-Max normalization [0, 1] on the training set to normalize the demand values. After training, we apply an inverse of the Min-Max transformation to recover the actual demand values. We use four ConvLSTM layers with 32/16/8/4 number of filters and 3×3 size of filters. We set the sequence length 24 hours for ConvLSTM. We use the Batch normalization after each ConvLSTM layer. The batch size is 32 and number of epochs are 500. We choose first 90% of the data as training data and the remaining 10% is used for the testing. We apply Leaky ReLU for fully-connected layers as the activation function. We use the stochastic gradient descent (SGD) algorithm with learning rate 0.0001. Furthermore, to prevent overfitting, we apply the dropout method with probability 0.4. Our model is implemented with Keras, a Deep Learning Python library [17] and is trained on two nVidia P100 GPUs.

*E. Performance Comparison*

**Comparison with state-of-the-art methods.** Fig. 3 shows the preprocessing on the real data and the results obtained by D-GAN. We model the Taxi demand in the transportation network into matrices analogous to the pixel matrices of an image, named ST map. The color map shows the number of taxi requests for a grid, dark means high number of requests.

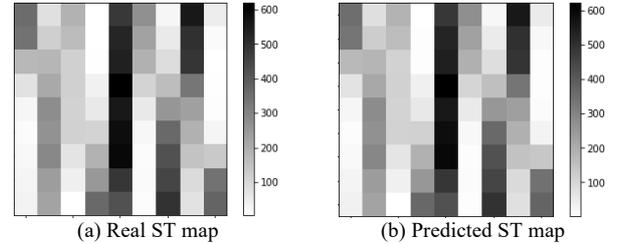

(a) Real ST map  (b) Predicted ST map

Fig. 3. Visualization result of the predicted map by the D-GAN for Taxi dataset

Table I shows the performance of D-GAN in comparison to other baselines methods for two datasets. D-GAN performs best among other approaches with lowest 1.47 and 0.10 RMSE, and 1.04 and 0.07 MAE for Taxi and Bike datasets, respectively. It achieves 45.95% and 47.36% (RMSE) relative improvement over the best performance among all baseline methods for Taxi and Bike datasets, respectively (see Fig. 4 (a) and (d)).

Traditional statistical approaches Moving Average, Weighted Moving Average and ARIMA perform quite poor with RMSE of 11.41, 10.88, and 11.35, respectively, as these models predict demand using average values of previous demands given in the same relative time interval. Regression methods, such as Linear Regression (LR), Random Forest (RF), XGBoost, and MLP achieve better performance than previous ones but still these methods are not able to capture

ST correlations in the ST data. We further extend our model comparison with the well-known deep learning methods, such as LSTM and CNN. LSTM cannot model spatial correlations while CNN cannot capture temporal dynamics. To model ST relationships, ST-ResNet and DMVST-Net are the state-of-the-art approaches for grid-based data prediction. Both approaches achieve improvements over the LSTM and CNN but still results show limited performance as result obtained through traditional machine learning model uses mean squared error (MSE) which is a pixel-wise average over many slightly different possible solutions in the pixel space [9]. Our proposed model learns rich distributions implicitly over data which is hard to model with an explicit likelihood.

As we have already discussed, GAN has shown its potential and has been widely used in many applications, such as gaming, image and video generation, visual computing, etc. Therefore, we train the LSTM and CNN models (Tensor size (TS): (9, 9)) in an adversarial way. Results show that GAN-based models are performing better than existing deep learning models as they are learning the sample generation from the probability distribution. Even though, basic GAN based LSTM and CNN models are performing better than LSTM and CNN but basic GAN with input noise cannot adequately cover the space of possible futures.

TABLE I. PERFORMANCE COMPARISON FOR THE NEW YORK CITY – TAXI DATASET AND BIKE DATASET

| Methods | RMSE_Taxi | MAE_Taxi | RMSE_Bike | MAE_Bike |
|---|---|---|---|---|
| Simple Moving Average | 11.41 | 11.05 | 10.11 | 9.33 |
| Weighted Moving Average | 10.88 | 10.12 | 9.56 | 9.04 |
| ARIMA | 11.35 | 10.98 | 10.86 | 10.43 |
| LR | 6.65 | 4.74 | 6.44 | 5.15 |
| RF | 5.97 | 4.20 | 5.03 | 4.78 |
| XGBOOST | 5.85 | 4.29 | 4.98 | 4.37 |
| MLP | 5.21 | 2.83 | 4.78 | 2.01 |
| LSTM | 3.81 | 1.39 | 4.22 | 1.97 |
| CNN | 3.44 | 1.28 | 3.86 | 1.82 |
| ST-Res-Net | 3.38 | 1.74 | 0.46 | 0.41 |
| DMVST-Net | 3.30 | 1.49 | 0.42 | 0.38 |
| G-LSTM | 2.72 | 1.63 | 0.19 | 0.14 |
| G-CNN | 3.27 | 2.45 | 0.35 | 0.30 |
| G-ConvLSTM | 4.68 | 3.93 | 0.29 | 0.23 |
| **D-GAN (Ours)** | **1.47** | **1.04** | **0.10** | **0.07** |

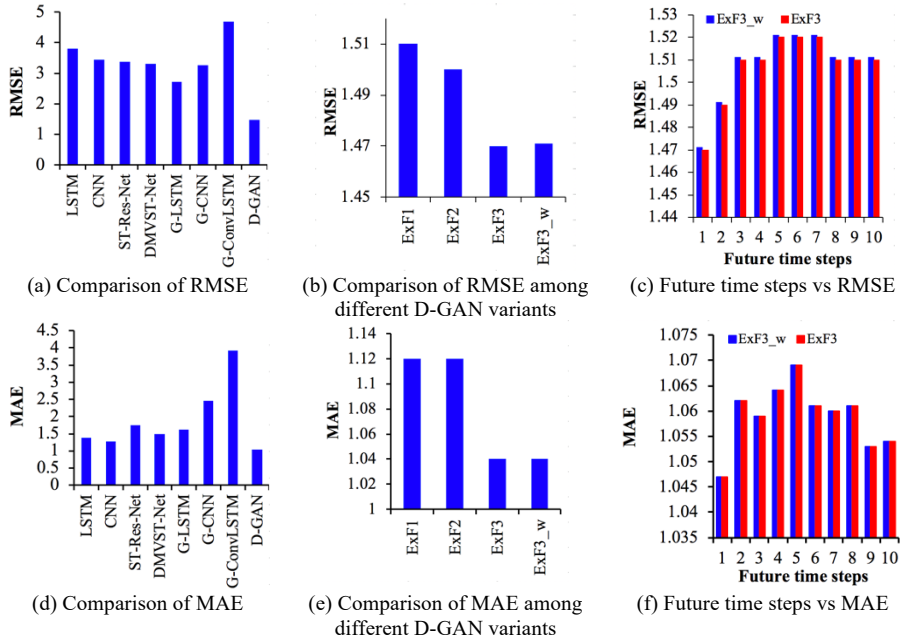

(a) Comparison of RMSE  
(b) Comparison of RMSE among different D-GAN variants  
(c) Future time steps vs RMSE  
(d) Comparison of MAE  
(e) Comparison of MAE among different D-GAN variants  
(f) Future time steps vs MAE  

Fig. 4. Evaluation of D-GAN network for the NYC Taxi dataset

We then implement G-ConvLSTM (TS: (24, 9, 9, 1)) where each input feature of a ConvLSTM network is a three-dimensional ST tensor. Studies show that ConvLSTM network captures better ST correlations and consistently outperforms LSTM [13]. ConvLSTM's results show that accuracy is less in compared to two-dimensional input of data, G-LSTM and G-CNN for the Taxi dataset (see Fig. 4 (a) and (d)). The possible reason can be that basic GAN cannot capture ST correlations for the high dimensional data in comparison to low dimensional data. We further extend our analysis and combined the VAE with GAN (using ConvLSTM), named D-GAN for ST prediction as features from different spaces are complementary for capturing better ST relationships. Results in Table I show that D-GAN achieves more accurate ST prediction in comparison to both traditional and deep learning based ST prediction models as it can handle the highly-structured yet stochastic nature of ST data.

TABLE II. D-GAN AND ITS VARIANTS

| Methods | RMSE_Taxi | MAE_Taxi | RMSE_Bike | MAE_Bike |
|---|---|---|---|---|
| D-GAN-LS | 8.89 | 7.65 | 7.73 | 6.86 |
| ExF1 | 1.51 | 1.12 | 0.172 | 0.12 |
| ExF2 | 1.5 | 1.12 | 0.17 | 0.12 |
| **ExF3** | **1.47** | **1.04** | **0.10** | **0.07** |
| ExF3_w | 1.472 | 1.04 | 0.101 | 0.07 |

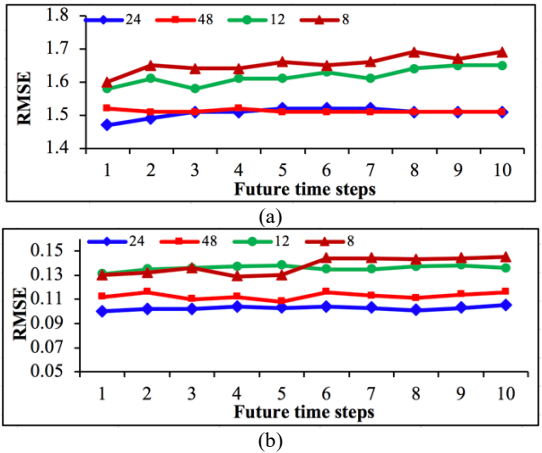

Fig. 5. Impact of sequence length on RMSE and future time steps for (a) NYC taxi dataset and (b) NYC Bike dataset

**Comparison with variants of D-GAN.** We also study the effect of different variants of our proposed model. Table II shows the performance comparison of D-GAN with other variants.

- **Loss Selection:** To show the importance of the reconstruction loss, we use the feature similarity loss instead of L2 loss (D-GAN-LS). The result shows L2 loss performs better for the ST data. This may be possible as two ST maps can have small feature loss even if they are significantly different in pixel-by-pixel comparison.

- **Impact of External Factors:** D-GAN model combines PoI data, weekday/weekend (dw) and weather data (w) as external factors. D-GAN with PoI data, with PoI and dw, and with PoI, dw and w are represented as ExF1, ExF2 and ExF3, respectively. ExF3 performs better than ExF1 and approximately equivalent to D-GAN without any external factor (ExF3_w) as shown in Fig. 4 (b) and (e).

- **Multi-step Prediction:** We further extend our analysis to find the impact of multi-step prediction on the model's accuracy. We predict ST maps for 10 future steps from one ST map and notice that the accuracy difference between these two models remain approximately same (see Fig. 4 (c) and (f)). The possible reason is that D-GAN implicitly learns the semantic variations and ST correlations and underlying factors of variations in the data.

- **Impact of Sequence Length:** We also study how the sequence length of ST maps affects the model's performance. Fig. 5 shows the prediction error of RMSE w.r.to the sequence length for both datasets. We observe that for sequence length of 24 hours, our model achieves the best performance for both datasets. We test the performance for 8 and 12 hours where RMSE degraded. The possible reason can be that model could not capture temporal dependencies. As the sequence length increases to 48, at the step 1, the prediction error degrades slightly while in later steps, performance is approximately equal to the performance of 24 hours. The possible reason is that calculating longer temporal dependency means to train higher number of parameters, i.e., training becomes hard.

VI. RELATED WORK

In this section, we shall discuss the related works on ST prediction.

**Time-Series Prediction.** In recent times, many effective statistics and deep learning based models have been proposed for ST prediction. Autoregressive integrated moving average (ARIMA) and its variants have been widely applied for ST prediction [18][19][20]. But these models are not able to capture spatial and temporal relations. [20] proposed a framework where predicted demand is a weighted ensemble of three prediction models. While, some researchers aim to predict travel speed and traffic volume on the road [21][22]. These methods predict traffic volume only for single or multiple road segments instead of for a city. Then time-series based prediction approaches [23][24] are proposed to capture the spatial relations with the external context data, such as weather, holiday, etc. But still these models could not model the complex non-linear ST relationship. On the other hand, traditional works can predict only for particular region at a time instead of city-level ST prediction.

**Deep Learning for ST Data.** Besides traditional time-series models, deep learning based models achieved a great success for the modeling of ST data. Some researchers used context data from multiple sources and modeled that data using a stack of several fully connected layers for traffic demand prediction [25], and taxi supply-demand gap [26]. These models do not consider the spatial and temporal relations explicitly.

Some researchers explored the CNN to capture spatial correlation for ST prediction [2][5]. On the other hand, some researchers used RNN to model temporal dependencies in the data [7]. Modeling spatial dependency and temporal dynamics separately could not capture ST correlations in the data.

A new solution, convolutional LSTM [13] is proposed to model the spatial and temporal relationships together for precipitation nowcasting. ST-ResNet [2] modelled temporal closeness, trend, and period using the residual neural network for predicting the crowd flow in a city. Recently, [8] proposed ST network for predicting demand, and considered ST temporal correlations and semantic variations. This network used local CNN, LSTM and graph embedding for the spatial, temporal and semantic views, respectively. However, existing deep-learning based ST prediction models are trained using mean square error *by constructions* which cannot model multi-modal data [9][27].

In summary, D-GAN is based on GAN architecture which can work with multi-modal outputs and can successfully handle highly-structured and stochastic nature of ST data. D-GAN learns to predict future ST map from a sequence of ST maps through learning construction of non-trivial internal representation which allows D-GAN to learn ST correlations, ST semantic variations and underlying factors of variations in the data simultaneously.

## VII. Conclusion

In this paper, we propose a novel deep generative adversarial based network (named, D-GAN) for more accurate ST data prediction. To the best of our knowledge, this is the first work to extract ST correlations and capture variations in the data for ST prediction via unsupervised and deep implicit generative model. D-GAN is inspired by GAN structure which is a structured probabilistic model that consists of two adversarial models: a generative model (G) to capture the data distribution and a discriminative model (D) to estimate the probability that a sample is generated from the training data or generated by the G. The two players, Discriminator (D) and Generator (G) play a two-player minimax game until Nash equilibrium. However, GAN model with input noise does not generate diverse future. Therefore, D-GAN integrates the directed graphical model, i.e., VAE with GAN into an unsupervised generative model. D-GAN successfully handles the ST correlations and stochastic elements that exists within ST data. We also design fusion module to incorporate the external factors information to improve the estimating capability of *G*. In addition, *D* learns to discriminate jointly in data and latent space for better learning and stable training. We evaluated the performance of our proposed model on a case study of demand prediction using two real-world datasets. The results show that D-GAN is achieving more accurate performance than 14 mainstream baseline methods. In the future, we will try to use attention mechanism to improve the model's performance.